\documentstyle[aclap,psfig]{article}

\title{A New Statistical Parser Based on Bigram Lexical Dependencies}
\author{Michael John Collins\thanks{This research was supported by ARPA 
Grant N6600194-C6043.}
\\ Dept. of Computer and Information Science \\
University of Pennsylvania \\ Philadelphia, PA, 19104, U.S.A. \\
{\tt mcollins@gradient.cis.upenn.edu}}

\begin{document}
\bibliographystyle{fullname}
\maketitle
\vspace{-0.5in}
\begin{abstract}
{This paper describes a new statistical parser which is based on
probabilities of dependencies between head-words in the parse tree.
Standard bigram probability estimation techniques are extended to calculate
probabilities of dependencies between pairs of words. Tests using
Wall Street Journal data show that the method performs at 
least as well as SPATTER \cite{magerman,ibm}, which has the best published
results for a statistical parser on this task. The simplicity of the approach 
means the model trains on 40,000 sentences in under 15 minutes. With a
beam search strategy parsing speed can be improved to over
200 sentences a minute with negligible loss in accuracy.}
\end{abstract}

\section{Introduction}

Lexical information has been shown to be
crucial for many parsing decisions, such as
prepositional-phrase attachment (for example \cite{hindle}).
However, early approaches to probabilistic parsing 
\cite{pereira,magerman and marcus,briscoe}
conditioned probabilities on non-terminal labels and part of speech tags alone.
The SPATTER parser \cite{magerman,ibm} does
use lexical information, and recovers labeled constituents in Wall
Street Journal text with above 84\% accuracy --
as far as we know the best published results on this task.

This paper describes a new parser which is much simpler 
than SPATTER, yet performs at least as well when trained and tested
on the same Wall Street Journal data.
The method uses lexical information directly by modeling 
head-modifier\footnote{By `modifier' we mean the linguistic
notion of either an argument or adjunct.}
relations between pairs of words. In this way it is similar to 
Link grammars \cite{lafferty}, and dependency grammars in general.

\section{The Statistical Model}

\begin{figure*}[htb]
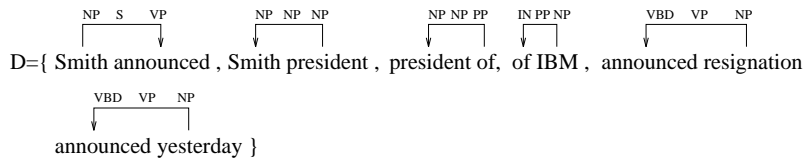


(a)

\begin{quote}
John/NNP Smith/NNP, the/DT president/NN of/IN IBM/NNP, announced/VBD his/PRP\$ resignation/NN yesterday/NN .
\end{quote}

\vspace{0.25in}

(b)

\centerline{\psfig{figure=wholetree2.eps,width=4.1in}}

\vspace{0.25in}

(c)

\centerline{\psfig{figure=newdepend2.eps,width=4.5in}}

\vspace{0.25in}

(d)

\centerline{\psfig{figure=newmandd2.eps,width=4.1in}}

\caption{An overview of the representation used by the model.
{\bf (a)} The tagged sentence; {\bf (b)} A candidate parse-tree (the correct one); 
{\bf (c)} A dependency representation of (b). Square
brackets enclose baseNPs (heads of baseNPs are marked in bold). 
Arrows show modifier $\rightarrow$ head dependencies.
Section 2.1 describes how arrows are labeled with non-terminal triples
from the parse-tree.
Non-head words within baseNPs are excluded from the
dependency structure; {\bf (d)} $B$, the set of baseNPs,
and $D$, the set of dependencies, are extracted from (c).}
\label{fig-overview}
\end{figure*}

The aim of a parser is to take a tagged sentence as input 
(for example Figure~\ref{fig-overview}(a))
and produce a phrase-structure tree as output (Figure~\ref{fig-overview}(b)).
A statistical approach to this problem consists of two components.
First, the {\em statistical model} assigns a probability to every candidate
parse tree for a sentence. Formally, given a sentence $S$ and a tree $T$, 
the model estimates the conditional probability $P(T|S)$. The most likely 
parse under the model is then:

\begin{equation}
T_{best} = argmax_T P(T|S)
\end{equation}

Second, the {\em parser} is a method for finding $T_{best}$.
This section describes the statistical model, while section~\ref{sec-parser}
describes the parser. 

The key to the statistical model is that any tree such as 
Figure~\ref{fig-overview}(b) can be represented as a set of 
{\bf baseNPs}\footnote{A baseNP or
`minimal' NP is a non-recursive NP, i.e. none of its
child constituents are NPs. The term was first used in \cite{ramshaw}.}
and a set of {\bf dependencies} as in Figure~\ref{fig-overview}(c). We call
the set of baseNPs ${\boldmath B}$, and the set of dependencies 
${\boldmath D}$;
Figure~\ref{fig-overview}(d) shows $B$ and $D$ for this example. For the purposes of our model, \mbox{$T=(B,D)$}, and:
\begin{equation}
P(T|S) = P(B,D|S) = P(B|S) \times P(D|S,B)
\label{eq-pts}
\end{equation}

$S$ is the sentence with words tagged for part of speech.
That is, $S=<(w_1,t_1),(w_2,t_2) ... (w_n,t_n)>$.
For POS tagging we use a maximum-entropy tagger described in
\cite{adwait}. The tagger performs at around 97\% accuracy on Wall
Street Journal Text, and is trained on the first 40,000 sentences of
the Penn Treebank \cite{marcus}. 

Given $S$ and $B$, the {\bf reduced sentence} $\bar{S}$ is defined as the 
subsequence of $S$ which is formed by removing punctuation and
reducing all baseNPs to their head-word alone. 

Thus the reduced sentence is an array of
word/tag pairs, \mbox{$\bar{S}=<\langle\bar{w}_1,\bar{t}_1\rangle,\langle\bar{w}_2,\bar{t}_2\rangle ... \langle\bar{w}_m,\bar{t}_m\rangle>$},
where $m \leq n$. For example for Figure~\ref{fig-overview}(a)
\newtheorem{sentence}{Example}
\begin{sentence}
$\bar{S} = \newline
\begin{array}{ll}
\;\;<&\langle Smith,NNP \rangle, \langle president,NN \rangle,\langle of,IN \rangle,\\
&\langle IBM,NNP \rangle, \langle announced,VBD \rangle,\\
&\langle resignation,NN \rangle, \langle yesterday,NN \rangle \;\; >\end{array}$
\label{sent2}
\end{sentence}

Sections 2.1 to 2.4 describe the dependency model.
Section 2.5 then describes the baseNP model, which uses bigram tagging
techniques similar to \cite{ramshaw,church}.

\subsection{The Mapping from Trees to Sets of Dependencies}
\label{sec-d}

The dependency model is limited to relationships between
words in  {\bf reduced} sentences such as Example~\ref{sent2}. 
The mapping from trees to dependency structures is central 
to the dependency model. It is defined in two steps:

{\bf 1.} For each constituent
\mbox{$P \rightarrow <C_1...C_n>$} in the parse tree 
a simple set of rules\footnote{The rules are essentially the same
as in \cite{magerman,ibm}. These rules are also used to 
find the head-word of baseNPs, enabling the mapping from $S$ and $B$
to $\bar{S}$.} 
identifies which of the children $C_i$ is the `head-child' of $P$.
For example, {\tt NN} would be identified as the head-child of 
\mbox{{\tt NP $\rightarrow$ $<$DET JJ JJ NN$>$}},
{\tt VP} would be identified as the head-child of 
\mbox{{\tt S $\rightarrow$ $<$NP VP$>$}}.
Head-words propagate up through the tree, each parent receiving
its head-word from its head-child. For example,
in \mbox{{\tt S $\rightarrow$ <NP VP>}},
{\tt S} gets its head-word, $announced$, from its head-child, the {\tt VP}.

\begin{figure}[h]
\centerline{\psfig{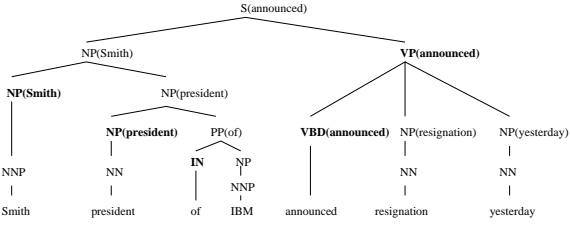}}
\caption{Parse tree for the reduced sentence in Example 1.
The head-{\bf child} of each constituent is shown in bold.
The head-{\bf word} for each constituent is shown in parentheses.}
\label{fig-howhead}
\end{figure}

\begin{figure}[h]
\centerline{\psfig{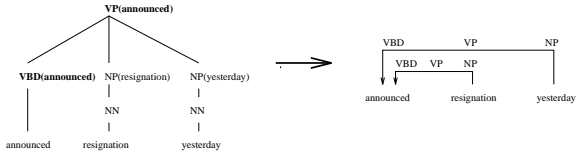}}
\caption{Each constituent with $n$ children (in this case $n=3$)
contributes $n-1$ dependencies.}
\label{fig-howdepend}
\end{figure}

{\bf 2.} Head-modifier relationships are now extracted
from the tree in Figure~\ref{fig-howhead}. Figure~\ref{fig-howdepend} 
illustrates how
each constituent contributes a set of dependency relationships.
{\tt VBD} is identified as the head-child of 
\mbox{{\tt VP $\rightarrow$ $<$VBD NP NP$>$}}. The head-words
of the two NPs, $resignation$ and $yesterday$, both modify the head-word
of the {\tt VBD}, $announced$. Dependencies are labeled by the modifier
non-terminal, {\tt NP} in both of these cases, the parent non-terminal, 
{\tt VP}, and finally the head-child non-terminal, {\tt VBD}.
The triple of non-terminals at the start, middle and end of the arrow
specify the nature of the dependency relationship -- {\tt $<$NP,S,VP$>$} 
represents a subject-verb dependency,
{\tt $<$PP,NP,NP$>$} denotes prepositional phrase modification of an {\tt NP},
and so on\footnote{The triple can also be viewed as representing a
semantic predicate-argument relationship, with the three elements being
the type of the  argument, result and functor respectively. This is
particularly apparent in Categorial Grammar formalisms \cite{cg}, which
make an explicit link between dependencies and functional application.}.

Each word in the reduced sentence, with the exception of the sentential head
`announced', modifies exactly one other word.
We use the notation \begin{equation}AF(j)=(h_j,R_j)\end{equation}
to state that the $j$th word in the reduced sentence
is a modifier to the ${h_j}$th word, with 
relationship $R_j$\footnote{For the head-word of the entire sentence
\hbox{$h_j=0$}, with $R_j$=$<$Label of the root of the parse tree $>$.
So in this case, $AF(5)=(0,<S>)$.}. $AF$ stands for `arrow from'.
$R_j$ is the triple of labels at the start, middle and end of 
the arrow. For example, $\bar{w}_1=Smith$ in this
sentence, and $\bar{w}_5=announced$, so \mbox{$AF(1)=(5,<${\tt NP,S,VP}$>)$}.

$D$ is now defined as the m-tuple of dependencies:
$D=\{(AF(1), AF(2) ... AF(m)\}$. The model assumes 
that the dependencies are independent, so that:

\begin{equation}
P(D|S,B)=\prod_{j=1}^{m}P(AF(j)|S,B)
\label{eq-pdsm}
\end{equation}


\subsection{Calculating Dependency Probabilities}

This section describes the way $P(AF(j)|S,B)$ is estimated. 
The same sentence is very unlikely to
appear both in training and test data, so we need to back-off from
the entire sentence context. We believe that
lexical information is crucial to attachment
decisions, so it is natural to condition on the words and tags. 
Let ${\cal V}$ be the vocabulary of all words seen in training data, ${\cal T}$
be the set of all part-of-speech tags, and ${\cal TRAIN}$ be the training
set, a set of reduced sentences. We define the following functions:

$\bullet$ $C\,(\, \langle a,b \rangle \, , \, \langle c,d \rangle \, )$ for
$a,c \;\epsilon\; {\cal V}$, and $b,d \;\epsilon\; {\cal T}$
is the number of times $\, \langle a,b \rangle \,$ and $\, \langle c,d \rangle \,$ 
are seen in the same reduced sentence in training data.\footnote{Note that
we count multiple co-occurrences in a single sentence, e.g.
if \hbox{$\bar{S}=(<a,b>,<c,d>,<c,d>)$} 
then \hbox{$C(<a,b>,<c,d>)=C(<c,d>,<a,b>)=2$}.}
Formally,
\begin{eqnarray}
C\,(\, \langle a,b \rangle \, , \, \langle c,d \rangle \, ) = & &\nonumber \\
& & \hspace{-1.6in} \sum_{
\stackrel{\bar{S} \;\epsilon\; {\cal TRAIN}}
{k,l = 1 .. |\bar{S}|, \;l \neq k}
}
\hspace{-0.2in}
h\left(\bar{S}[k]=\langle a,b \rangle \, , \, \bar{S}[l]=\langle c,d \rangle \right)
\end{eqnarray}
where $h(x)$ is an indicator function which is $1$ if $x$ is true,
$0$ if $x$ is false.

$\bullet$ $C\,(R,\, \langle a,b \rangle \, , \, \langle c,d \rangle \, )$
is the number of times $\, \langle a,b \rangle \,$ and $\, \langle c,d \rangle \,$ 
are seen in the same reduced sentence in training data,
and $\, \langle a,b \rangle \,$ modifies $\, \langle c,d \rangle \,$ 
with relationship $R$.
Formally,
\begin{eqnarray}
C\,(R,\, \langle a,b \rangle \, , \, \langle c,d \rangle \, ) = & &\nonumber \\
& & \hspace{-1.6in}\sum_{
\stackrel{\bar{S} \;\epsilon\; {\cal TRAIN}}
{k,l = 1 .. |\bar{S}|, \;l \neq k}
}
\hspace{-0.2in}
h(\bar{S}[k]=\langle a,b \rangle \, , \, \bar{S}[l]=\langle c,d \rangle \, , \,AF(k)=(l,R)) \nonumber\\
&&
\end{eqnarray}


$\bullet$ $F(R \, | \, \langle a,b \rangle \, , \, \langle c,d \rangle \,)$ is the probability that 
$\, \langle a,b \rangle \,$ modifies $\, \langle c,d \rangle \,$ with relationship $R$, given that
$\, \langle a,b \rangle \,$ and $\, \langle c,d \rangle \,$ appear in the same 
reduced sentence.
The maximum-likelihood estimate of $F(R \, | \, \langle a,b \rangle \, , \, \langle c,d \rangle \,)$ is:
\begin{equation}
\hat{F}(R \, | \, \langle a,b \rangle \, , \, \langle c,d \rangle \,) = \frac{C(R , \, \langle a,b \rangle \, , \, \langle c,d \rangle \,)}{C(\, \langle a,b \rangle \, , \, \langle c,d \rangle \,)}
\label{eq-ml}
\end{equation}
We can now make the following approximation:
\begin{eqnarray}
\lefteqn{P(AF(j)=(h_j,R_j) \; | \; S,B) \approx} \nonumber \\
& & \frac{\hat{F}(R_j \, |\, \langle \bar{w}_j,\bar{t}_j \rangle \,,\, \langle \bar{w}_{h_j},\bar{t}_{h_j} \rangle \,)}
{\sum_{k=1..m,k\neq j,p\epsilon {\cal P}}\hat{F}(p \, | \, \langle \bar{w}_j,\bar{t}_j \rangle \,,\, \langle \bar{w}_k,\bar{t}_k \rangle \,)}
\label{eq-mods}
\end{eqnarray}
where ${\cal P}$ is the set of all triples of non-terminals.
The denominator is a normalising factor which ensures that
\[\sum_{k=1..m,k\neq j, p\epsilon {\cal P}}P(AF(j)=(k,p) \; | \; S,B) = 1\]
From~(\ref{eq-pdsm})~and~(\ref{eq-mods}):
\begin{eqnarray}
\lefteqn{P(D|S,B) \approx} \label{eq-pdms2} \\
& & \prod_{j=1}^{m}\frac{\hat{F}(R_j \, |\, \langle \bar{w}_j,\bar{t}_j \rangle \,,\, \langle \bar{w}_{h_j},\bar{t}_{h_j} \rangle \,)}
{\sum_{k=1..m,k\neq j,p\epsilon {\cal P}}\hat{F}(p \, | \, \langle \bar{w}_j,\bar{t}_j \rangle \,,\, \langle \bar{w}_k,\bar{t}_k \rangle \,)} \nonumber
\end{eqnarray}
The denominator of~(\ref{eq-pdms2}) is constant, so
maximising $P(D|S,B)$ over $D$ for fixed $S,B$ is equivalent to
maximising the product of the numerators, ${\cal N}(D|S,B)$.
(This considerably simplifies the parsing process):

\begin{equation}
{\cal N}(D|S,B)=\prod_{j=1}^{m}\hat{F}(R_j \, |\, \langle \bar{w}_j,\bar{t}_j \rangle \,,\, \langle \bar{w}_{h_j},\bar{t}_{h_j} \rangle \,)
\label{eq-numer}
\end{equation}

\subsection{The Distance Measure}

An estimate based on the identities of the two tokens alone is problematic. 
Additional context, in particular the relative order of the two words and
the distance between them, will also 
strongly influence the likelihood of one word modifying the other.
For example consider the relationship between `sales' and the three tokens 
of `of':

\begin{sentence}
Shaw, based in Dalton, Ga., has annual {\bf sales of} about \$ 1.18
billion, and has economies {\bf of} scale and lower raw-material costs that are expected to boost the profitability {\bf of} Armstrong 's brands, sold under the Armstrong and Evans-Black names .
\end{sentence}

In this sentence `sales' and `of' co-occur three times. 
The parse tree in training data
indicates a relationship in only one of these cases, 
so this sentence would contribute an estimate
of $\frac{1}{3}$ that the two words are related. This seems unreasonably
low given that `sales of' is a strong collocation. 
The latter two instances of `of' are so distant from `sales' that it is
unlikely that there will be a dependency.

This suggests that
distance is a crucial variable when deciding whether two words are related.
It is included in the model by defining an extra 
`distance' variable, $\Delta$, and extending $C$, $F$ and $\hat{F}$ to
include this variable. For example, $C(\, \langle a,b \rangle \, , \, \langle c,d \rangle \,,\Delta)$ is
the number of times $\, \langle a,b \rangle \,$ and $\, \langle c,d \rangle \,$ appear in the same sentence
at a distance $\Delta$ apart.
(\ref{eq-finaln}) is then maximised instead of~(\ref{eq-numer}):
\begin{equation}
{\cal N}(D|S,B)=\prod_{j=1}^{m}\hat{F}(R_j \, |\, \langle \bar{w}_j,\bar{t}_j \rangle \,,\, \langle \bar{w}_{h_j},\bar{t}_{h_j} \rangle \,,\Delta_{j,h_j})
\label{eq-finaln}
\end{equation}
A simple example of $\Delta_{j,h_j}$ would be \hbox{$\Delta_{j,h_j}=h_j-j$}.
However, other features of a sentence, such as punctuation, are also useful
when deciding if two words are related. We have developed a 
heuristic `distance' measure which takes several such features into account
The current distance measure $\Delta_{j,h_j}$ is the combination of
6 features, or questions (we motivate the choice of these questions
qualitatively -- section~\ref{sec-discussion} gives quantitative results 
showing their merit):

{\bf Question 1} Does the $h_j$th word precede or follow the $j$th word?
English is a language with strong word order, so the 
order of the two words in surface text will clearly affect their 
dependency statistics.

{\bf Question 2} Are the $h_j$th word and the $j$th word adjacent?
English is largely right-branching and head-initial, which
leads to a large proportion of dependencies being between adjacent words
\footnote{For example in `(John (likes (to (go (to (University (of
 Pennsylvania)))))))' all dependencies are between adjacent words.}.
Table~\ref{table-dist} shows just how local most dependencies are.

\begin{table}[h]
\begin{center}
\begin{tabular}{lrrrr}
\hline
Distance&1&$\leq 2$&$\leq 5$&$\leq 10$\\
Percentage&74.2&86.3&95.6&99.0\\
\hline
\end{tabular}
\caption{Percentage of dependencies vs. distance between the head 
words involved. These figures count baseNPs as  a single word,
and are taken from WSJ training data.}
\label{table-dist}
\end{center}
\end{table}

\begin{table}[h]
\begin{center}
\begin{tabular}{lrrr}
\hline
Number of verbs&0&$<=$1&$<=$2\\
Percentage&94.1&98.1&99.3\\
\hline
\end{tabular}
\caption{Percentage of dependencies vs. number of verbs between the head
words involved.}
\label{table-verbdepend}
\end{center}
\end{table}

{\bf Question 3} Is there a verb between the $h_j$th word and the $j$th word?
Conditioning on the exact distance between two words
by making $\Delta_{j,h_j}=h_j-j$ leads to severe sparse data problems. 
But Table~\ref{table-dist} shows the need to make
finer distance distinctions than just whether two words are adjacent.
Consider the prepositions `to', `in' and `of'
in the following sentence:

\begin{sentence}
Oil stocks {\bf escaped} the brunt of Friday 's selling and several were
able to post gains , including Chevron , which {\bf rose} 5/8 {\bf to} 
66 3/8 {\bf in} Big Board composite trading {\bf of } 2.4 million shares .
\label{sent-rose}
\end{sentence}

The prepositions' main candidates for attachment 
would appear to be the previous verb, 
`rose', and the baseNP heads between each preposition and this verb. 
They are less likely to modify a more distant verb such as `escaped'.
Question 3 allows the parser to prefer modification of the
most recent verb -- effectively another, weaker preference for 
right-branching structures. Table~\ref{table-verbdepend} shows
that 94\% of dependencies do not cross a verb, giving empirical evidence
that question 3 is useful.

\newpage

{\bf Questions 4, 5 and 6}$\:$

\begin{itemize}

\item Are there 0, 1, 2, or more than 2 `commas' between the $h_j$th word and the $j$th word?
(All symbols tagged as a `,' or `:' are considered to be `commas').

\item Is there a `comma' immediately following the first of the $h_j$th word and the $j$th word?

\item Is there a `comma' immediately preceding the second of the $h_j$th word and the $j$th word?

\end{itemize}

People find that punctuation is extremely useful for identifying 
phrase structure, and the parser described here
also relies on it heavily. Commas are not considered
to be words or modifiers in the dependency model -- 
but they do give strong indications about the parse structure. 
Questions 4, 5 and 6 allow the parser to use this information.

\subsection{Sparse Data}
\label{sec-backoff}

The maximum likelihood estimator in (\ref{eq-ml}) is likely to be 
plagued by sparse data problems -- $C(\, \langle \bar{w}_j,\bar{t}_j \rangle \,,\, \langle \bar{w}_{h_j},\bar{t}_{h_j} \rangle \,,\Delta_{j,h_j})$ 
may be too low to give a reliable estimate, or worse still it may be zero 
leaving  the estimate undefined. \cite{collins} describes how a
backed-off estimation strategy is used for making prepositional 
phrase attachment decisions. The
idea is to back-off to estimates based on less context.
In this case, less context means looking at the POS tags
rather than the specific words. 

There are four estimates,
$E_1$, $E_2$, $E_3$ and $E_4$, based respectively on: 1) both words and
both tags; 2) $\bar{w}_j$ and the two POS tags; 3) $\bar{w}_{h_j}$ and the 
two POS tags; 4) the two POS tags alone.

\begin{equation}
\begin{array}{cccc}
E_1=\frac{\eta_1}{\delta_1}&E_2=\frac{\eta_2}{\delta_2}&E_3=\frac{\eta_3}{\delta_3}&E_4=\frac{\eta_4}{\delta_4}\\
\end{array}
\end{equation}
where\footnote{
\[C(\, \langle \bar{w}_j,\bar{t}_j \rangle \,,\, \langle \bar{t}_{h_j} \rangle \,,\Delta_{j,h_j}) =
\sum_{x \epsilon {\cal V}} C(\, \langle \bar{w}_j,\bar{t}_j \rangle \,,\, \langle x,\bar{t}_{h_j} \rangle \,,\Delta_{j,h_j})\;\]
\[C(\, \langle \bar{t}_j \rangle \,,\, \langle \bar{t}_{h_j} \rangle \,,\Delta_{j,h_j}) =
\sum_{x \epsilon {\cal V}} \sum_{y \epsilon {\cal V}}C(\, \langle x,\bar{t}_j \rangle \,,\, \langle y,\bar{t}_{h_j} \rangle \,,\Delta_{j,h_j})\]
where ${\cal V}$ is the set of all words seen in training data: 
the other definitions of $C$ follow similarly.}
\begin{eqnarray}
\delta_1 & = & C(\, \langle \bar{w}_j,\bar{t}_j \rangle \,,\, \langle \bar{w}_{h_j},\bar{t}_{h_j} \rangle \,,\Delta_{j,h_j}) \nonumber 
\\ \vspace{0.4in}
\delta_2 & = & C(\, \langle \bar{w}_j,\bar{t}_j \rangle \,,\, \langle \bar{t}_{h_j} \rangle \,,\Delta_{j,h_j}) \nonumber 
\\ \vspace{0.4in}
\delta_3 & = & C(\, \langle \bar{t}_j \rangle \,,\, \langle \bar{w}_{h_j},\bar{t}_{h_j} \rangle \,,\Delta_{j,h_j}) \nonumber 
\\ \vspace{0.4in}
\delta_4 & = & C(\, \langle \bar{t}_j \rangle \,,\, \langle \bar{t}_{h_j} \rangle \,,\Delta_{j,h_j})   \nonumber 
\\ \vspace{0.4in}
\eta_1 &=&  C(R_j,\, \langle \bar{w}_j,\bar{t}_j \rangle \,,\, \langle \bar{w}_{h_j},\bar{t}_{h_j} \rangle \,,\Delta_{j,h_j})  \nonumber 
\\ \vspace{0.4in}
\eta_2 &=& C(R_j,\, \langle \bar{w}_j,\bar{t}_j \rangle \,,\, \langle \bar{t}_{h_j} \rangle \,,\Delta_{j,h_j})  \nonumber 
\\ \vspace{0.4in}
\eta_3 &=& C(R_j,\, \langle \bar{t}_j \rangle \,,\, \langle \bar{w}_{h_j},\bar{t}_{h_j} \rangle \,,\Delta_{j,h_j})  \nonumber 
\\ \vspace{0.4in}
\eta_4 &=& C(R_j,\, \langle \bar{t}_j \rangle \,,\, \langle \bar{t}_{h_j} \rangle \,,\Delta_{j,h_j})
\label{eq-counts}
\end{eqnarray}

Estimates 2 and 3 compete -- for a given pair of words in test data both 
estimates may exist and they are equally `specific' to the
test case example. \cite{collins} suggests the
following way of combining them, which favours the estimate
appearing more often in training data:

\begin{equation}
E_{23} = \frac{\eta_2 + \eta_3}{\delta_2 + \delta_3}
\end{equation}

This gives three estimates: $E_1$, $E_{23}$ and $E_4$, a similar situation
to trigram language modeling for speech recognition \cite{jelinek}, 
where there are trigram, bigram and unigram estimates.
\cite{jelinek} describes a deleted interpolation method which
combines these estimates to give a `smooth' estimate,
and the model uses a variation of this idea:

\vspace{1ex}

{\bf If ${\boldmath E_1}$ exists}, i.e. $\delta_1 > 0$
\begin{eqnarray}
\lefteqn{\hat{F}(R_j \, |\, \langle \bar{w}_j,\bar{t}_j \rangle \,,\, \langle \bar{w}_{h_j},\bar{t}_{h_j} \rangle \,,\Delta_{j,h_j}) =} \nonumber \\
& & \lambda_1 \times E_1 + (1 - \lambda_1) \times E_{23}
\label{eq-est1}
\end{eqnarray}

{\bf Else If ${\boldmath E_{23}}$ exists}, i.e. $\delta_2 + \delta_3 > 0$
\begin{eqnarray}
\lefteqn{\hat{F}(R_j \, |\, \langle \bar{w}_j,\bar{t}_j \rangle \,,\, \langle \bar{w}_{h_j},\bar{t}_{h_j} \rangle \,,\Delta_{j,h_j}) = } \nonumber \\
& & \lambda_2 \times E_{23} + (1 - \lambda_2) \times E_4
\label{eq-est2}
\end{eqnarray}

{\bf Else}
\begin{eqnarray}
\hat{F}(R_j \, |\, \langle \bar{w}_j,\bar{t}_j \rangle \,,\, \langle \bar{w}_{h_j},\bar{t}_{h_j} \rangle \,,\Delta_{j,h_j})&=&E_4
\end{eqnarray}

\vspace{1ex}

\cite{jelinek} describes how to find $\lambda$ values 
in~(\ref{eq-est1})~and~(\ref{eq-est2}) which maximise the 
likelihood of held-out data. We have taken a simpler approach, namely:
\begin{eqnarray}
\lambda_1 & = & \frac{\delta_1}{\delta_1 + 1}\nonumber \\
&&\nonumber\\
\lambda_2 & = & \frac{\delta_2 + \delta_3}{\delta_2 + \delta_3 + 1}
\label{eq-l}
\end{eqnarray}
These $\lambda$ values have the desired property of increasing as the 
denominator of the more `specific' estimator increases. We think that a proper 
implementation of deleted interpolation is likely to improve results,
although basing estimates on co-occurrence counts alone has the advantage
of reduced training times.

\subsection{The BaseNP Model}
\label{sec-min}

The overall model 
would be simpler if we could do without the baseNP model and frame 
everything in terms of dependencies.
However the baseNP model is needed for two reasons. First, while
adjacency between
words is a good indicator of whether there is some relationship between them,
this indicator is made substantially stronger if baseNPs are reduced
to a single word. Second, it means that words internal to
baseNPs are not included in the co-occurrence counts in training data.
Otherwise, in a phrase like `The Securities and Exchange Commission closed
yesterday', pre-modifying nouns like `Securities' and `Exchange' would be
included in co-occurrence counts, when in practice there is no way that they
can modify words outside their baseNP.

The baseNP model can be viewed as tagging the gaps between words with 
$S(tart)$, $C(ontinue)$, $E(nd)$, $B(etween)$ or $N(ull)$ symbols, 
respectively meaning
that the gap is at the start of a $BaseNP$, continues a $BaseNP$, is
at the end of a $BaseNP$, is between two adjacent $baseNP$s, or is between two words
which are both not in $BaseNPs$.
We call the gap before the $i$th word $G_i$ (a sentence with $n$
words has $n-1$ gaps).
For example,  
{\quote [ John Smith ] [ the president ] of [ IBM ] has announced
[ his resignation ] [ yesterday ]}
$\Rightarrow$
{\quote John {\bf C} Smith {\bf B} the {\bf C} president {\bf E} of {\bf S} IBM {\bf E} has {\bf N} announced
{\bf S} his {\bf C} resignation {\bf B} yesterday}

The baseNP model considers the
words directly to the left and right of each gap, and whether there is
a comma between the two words (we write $c_i=1$ if there is a comma,
$c_i=0$ otherwise). Probability estimates are 
based on counts of consecutive pairs of words in {\bf unreduced} training 
data sentences, where baseNP boundaries 
define whether gaps fall into the $S$, $C$, $E$, $B$ or $N$ categories.
The probability of a baseNP sequence in an unreduced sentence $S$ is then:
\begin{equation}
\prod_{i=2 ... n}\hat{P}(G_i \; | \; w_{i-1},t_{i-1},w_{i},t_{i},c_i)\nonumber\\
\label{eq-basenp}
\end{equation}
The estimation method is analogous to that described in 
the sparse data section of this paper. The method is similar to that described
in \cite{ramshaw,church}, where baseNP detection is also framed as a tagging
problem.

\subsection{Summary of the Model}

The probability of a parse tree $T$, given a sentence $S$, is:
\[P(T|S) = P(B,D|S) = P(B|S) \times P(D|S,B)\]

The denominator 
in Equation~(\ref{eq-pdms2}) 
is not actually constant for different baseNP
sequences, but we make this approximation for the sake of
efficiency and simplicity. In practice this is a good approximation because
most baseNP boundaries are very well
defined, so parses which have high enough $P(B|S)$ to be among the highest
scoring parses for a sentence tend to have identical or very similar baseNPs.
Parses are ranked by the following 
quantity\footnote{In fact we also model the set of
unary productions, $U$, in the tree,
which are of the form $P \rightarrow <C_1>$. This introduces an additional
term, $\hat{P}(U|B,S)$, into~(\ref{eq-finalrank}).}:
\begin{equation}
\hat{P}(B|S) \times {\cal N}(D|S,B)
\label{eq-finalrank}
\end{equation}
Equations~(\ref{eq-basenp})~and~(\ref{eq-finaln}) define $\hat{P}(B|S)$ and 
${\cal N}(D|S,B)$. The parser finds the tree which 
maximises~(\ref{eq-finalrank}) subject to the hard constraint that 
dependencies cannot cross.

\subsection{Some Further Improvements to the Model}
\label{recent}

This section describes two modifications which improve
the model's performance.

$\bullet$ In addition to conditioning on whether dependencies cross commas,
a single constraint concerning punctuation is introduced.
If for any constituent {\tt Z} in the chart 
\mbox{{\tt Z $\rightarrow$ <.. X Y ..>}} two of its
children {\tt X} and {\tt Y} are
separated by a comma, then the last word in {\tt Y}
must be directly followed by a comma, or must be the last word in
the sentence. In training data 96\% of commas follow this rule. 
The rule also has the benefit of improving efficiency by reducing the number of
constituents in the chart.

$\bullet$ The model we have described thus far takes the single best sequence
of tags from the tagger, and it is clear that there is potential for better 
integration of the tagger and parser. We have tried two modifications.
First, the current estimation methods treat occurrences of the same
word with different POS tags as effectively distinct types. 
Tags can be ignored when lexical information is available by
defining \begin{equation}C\,(a,c) =
\sum_{b , d \epsilon {\cal T}} C\,(\, \langle a,b \rangle \, , \, \langle c,d \rangle \, )\end{equation} where ${\cal T}$ is the set of all tags. 
Hence $C\,(a,c)$ is the number of times that the words $a$ and $c$ occur in
the same sentence, ignoring their tags. The other definitions
in~(\ref{eq-counts}) are similarly redefined, with POS tags only being used
when backing off from lexical information. This makes the parser less sensitive
to tagging errors.

Second, for each word $w_i$ the tagger can provide
the distribution of tag probabilities $P(t_i|S)$ 
(given the previous two words are tagged as in the best overall sequence of
tags) rather than just the first best tag. The score for a parse 
in equation~(\ref{eq-finalrank}) then
has an additional term, $\prod_{i=1}^n P(t_i|S)$,
the product of probabilities of the tags which it contains.

Ideally we would like to integrate POS tagging into the parsing model rather 
than treating it as a separate stage. This is an area for future research.

\setlength{\dbltextfloatsep}{0ex}
\setlength{\textfloatsep}{2.5ex}

\begin{table*}[htb]
\begin{center}
\begin{tabular}{|c||c|c|c|c|c||c|c|c|c|c|}
\hline
MODEL&\multicolumn{5}{|c||}{$\leq$ 40 Words (2245 sentences)}&\multicolumn{5}{|c|}{$\leq$ 100 Words (2416 sentences)}\\ \cline{2-11}
&LR&LP&CBs&$0$ CBs&$\leq 2$ CBs&LR&LP&CBs&$0$ CBs&$\leq 2$ CBs\\
\hline
\hline
(1)&84.9\%&84.9\%&1.32&57.2\%&80.8\%&84.3\%&84.3\%&1.53&54.7\%&77.8\%\\
(2)&85.4\%&85.5\%&1.21&58.4\%&82.4\%&84.8\%&84.8\%&1.41&55.9\%&79.4\%\\
(3)&85.5\%&85.7\%&1.19&59.5\%&82.6\%&85.0\%&85.1\%&1.39&56.8\%&79.6\%\\
(4)&85.8\%&86.3\%&1.14&59.9\%&83.6\%&85.3\%&85.7\%&1.32&57.2\%&80.8\%\\
\hline
SPATTER&84.6\%&84.9\%&1.26&56.6\%&81.4\%&84.0\%&84.3\%&1.46&54.0\%&78.8\%\\
\hline
\end{tabular}
\caption{Results on Section 23 of the WSJ Treebank. {\bf (1)} is the basic 
model; {\bf (2)} is the basic model with the punctuation rule described in 
section 2.7;
{\bf (3)} is model (2) with POS tags ignored when lexical information 
is present;
{\bf (4)} is model (3) with probability distributions from the POS tagger.
{\bf LR/LP} = labeled
recall/precision. {\bf CBs} is the average number of crossing brackets per
sentence. {\bf 0 CBs, $\leq 2$ CBs} are the percentage of sentences with 0
or $\leq 2$ crossing brackets respectively.}
\label{tab-results}
\end{center}
\end{table*}

\section{The Parsing Algorithm}
\label{sec-parser}

\begin{figure}[htb]


\centerline{\psfig{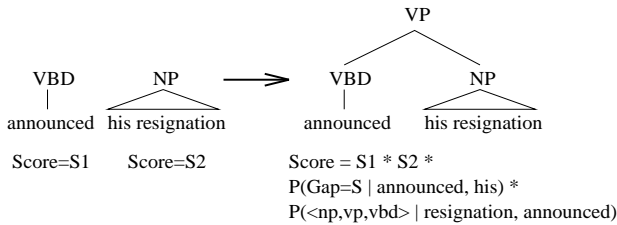}}



\caption{Diagram showing how two constituents join to form a new
constituent. Each operation gives two new probability terms: one for
the baseNP gap tag between the two constituents, 
and the other for the dependency between the head
words of the two constituents.
}
\label{fig-parsing}
\end{figure}

The parsing algorithm is a simple bottom-up chart parser. There is no
grammar as such, although in practice any dependency with a triple
of non-terminals which has not been seen in training data will get
zero probability. Thus the parser searches through the space of all
trees with non-terminal triples seen in training data. Probabilities of 
baseNPs in the chart are calculated using~(\ref{eq-basenp}), while 
probabilities for other 
constituents are derived from the dependencies and baseNPs that they contain.
A dynamic programming algorithm is used: if two proposed constituents span 
the same set of words, have the same label, head, and distance from the
head to the left and right end of the constituent, then the lower
probability constituent can be safely discarded. Figure~\ref{fig-parsing}
shows how constituents in the chart combine in a bottom-up manner.

\section{Results}
\label{sec-discussion}

The parser was trained on sections~02~-~21 of the Wall Street Journal 
portion of the Penn Treebank \cite{marcus}
(approximately 40,000 sentences), and tested on 
section~23 (2,416 sentences). For comparison SPATTER 
\cite{magerman,ibm}
was also tested on section~23. We use the PARSEVAL measures 
\cite{black} to compare performance:

\begin{description}

\item[Labeled Precision =]\begin{Large}
$\frac{number \: of \: correct \: constituents \: in \: proposed \: parse}
{number \: of \: constituents \: in \: proposed \: parse}$ \end{Large}

\item[Labeled Recall =]\begin{Large}
$\frac{number \:  of \:  correct \:  constituents \:  in \:  proposed \:  parse}
{number \:  of \:  constituents \:  in \:  treebank \:  parse}$ \end{Large}

\item[Crossing Brackets =] number of constituents which violate constituent 
boundaries with a constituent in the treebank parse.

\end{description}

For a constituent to be `correct' it must span the same set of 
words (ignoring punctuation, i.e. all tokens tagged as commas, colons
or quotes) and have the same label\footnote{SPATTER collapses {\tt ADVP} and
{\tt PRT} to the same label, for comparison
we also removed this distinction when calculating scores.} 
as a constituent in the treebank parse.
Four configurations of the parser were tested: {\bf (1)} The basic model;
{\bf (2)} The basic model with the punctuation rule described in
section~\ref{recent}; {\bf (3)} Model (2) with tags ignored when lexical
information is present, as described in~\ref{recent}; and {\bf (4)} Model
(3) also using the full probability distributions for POS tags.
We should emphasise that test data outside of section 23
was used for all development of the model,
avoiding the danger of implicit training on section 23.
Table~\ref{tab-results} shows the results of the tests.
Table~\ref{tab-discuss} shows results which indicate how different
parts of the system contribute to performance.

\begin{table}
\begin{center}
\begin{tabular}{|c|c|c|c|c|}
\hline
Distance&Lexical&LR&LP&CBs\\
Measure&Information&&&\\
\hline
\hline
Yes&Yes&85.0\%&85.1\%&1.39\\
Yes&No&76.1\%&76.6\%&2.26\\
No&Yes&80.9\%&83.6\%&1.51\\
\hline
\end{tabular}
\caption{The contribution of various components of the model.
The results are for all sentences of $\leq 100$ words in section 23
using model (3). For `no lexical information' all estimates are based on POS
tags alone. For `no distance measure' the distance measure is Question 1
alone (i.e. whether $\bar{w}_j$ precedes or follows $\bar{w}_{h_j}$).}
\label{tab-discuss}
\end{center}
\end{table}

\subsection{Performance Issues}

All tests were made on a Sun SPARCServer 1000E, using 100\% of a
60Mhz SuperSPARC processor.
The parser uses around 180 megabytes of memory, and
training on 40,000 sentences (essentially extracting the co-occurrence
counts from the corpus) takes under 15 minutes.
Loading the hash table of bigram counts into memory takes approximately
8 minutes.

Two strategies are employed to improve parsing efficiency. First,
a constant probability threshold is used while building the 
chart -- any constituents
with lower probability than this threshold are discarded. If a parse is
found, it must be the highest ranked parse by the model (as all constituents
discarded have lower probabilities than this parse and could not, therefore,
be part of a higher probability parse). If no parse is found, the threshold
is lowered and parsing is attempted again. The process continues until a parse
is found.

Second, a beam search strategy is used. For each span of words in the sentence
the probability, $P_h$, of the highest probability constituent is recorded.
All other constituents spanning the same words must
have probability greater than $\frac{P_h}{\beta}$ for some constant beam size
$\beta$ -- constituents 
which fall out of this beam are discarded. The method risks introducing
search-errors, but in practice efficiency can be greatly improved with
virtually no loss of accuracy. Table~\ref{tab-speed} 
shows the trade-off between speed and accuracy as the beam is narrowed.

\begin{table}[h]
\begin{center}
\begin{tabular}{|c|c|c|c|c|}
\hline
Beam&Speed&LR&LP&CBs\\
Size $\beta$&Sentences/minute&&&\\
\hline
\hline
1000&118&84.9\%&85.1\%&1.39\\
150&166&84.8\%&85.1\%&1.38\\
20&217&84.7\%&85.0\%&1.40\\
3&261&84.1\%&84.5\%&1.44\\
1.5&283&83.7\%&84.1\%&1.48\\
1.2&289&83.5\%&83.9\%&1.50\\
\hline
\end{tabular}
\caption{The trade-off between speed and accuracy as the beam-size is
varied. Model (3) was used for this test on all sentences $\leq 100$ words
in section 23.}
\label{tab-speed}
\end{center}
\end{table}

\section{Conclusions and Future Work}

We have shown that a simple statistical
model based on dependencies between words can parse Wall Street Journal 
news text with high accuracy. The method is equally applicable
to tree or dependency representations of syntactic structures.

There are many possibilities for improvement, which is encouraging.
More sophisticated estimation techniques such as deleted interpolation
should be tried.
Estimates based on relaxing the distance measure could also be used for
smoothing -- at present we only back-off on words.
The distance measure could be extended to capture more context,
such as other words or tags in the sentence. 
Finally, the model makes no account of valency.

\section*{Acknowledgements}

I would like to thank Mitch Marcus,
Jason Eisner, Dan Melamed and Adwait Ratnaparkhi for many useful discussions, 
and for comments on earlier versions of this paper. I would also like to
thank David Magerman for his help with testing SPATTER.

\end{document}